\documentclass{article}

\usepackage{microtype}
\usepackage{graphicx}
\usepackage{subfigure}
\usepackage{booktabs} % for professional tables

\usepackage{hyperref}

\usepackage[accepted]{icml2024}

\usepackage{amsmath}
\usepackage{amssymb}
\usepackage{mathtools}
\usepackage{amsthm}

\usepackage[capitalize,noabbrev]{cleveref}

\theoremstyle{plain}

\theoremstyle{definition}

\theoremstyle{remark}

\usepackage[textsize=tiny]{todonotes}

\usepackage{inconsolata}
\usepackage{multirow}
\usepackage{float}
\usepackage{amsfonts}
\usepackage{url}
\usepackage{bm}
\usepackage{xspace}
\usepackage{fdsymbol}
\usepackage{xcolor}
\usepackage{adjustbox}
\usepackage{pifont}
\usepackage{color, colortbl}
\definecolor{amber}{rgb}{1.0, 0.75, 0.0}
\definecolor{applegreen}{rgb}{0.55, 0.71, 0.0}
\definecolor{LightCyan}{rgb}{0.88,1,1}
\newcommand{\hlcell}{\cellcolor{applegreen!22}}

\usepackage[]{mdframed}
\usepackage{tikz}
\newcommand*\circled[1]{\tikz[baseline=(char.base)]{
            \node[shape=circle,draw,inner sep=1.0pt] (char) {#1};}}
\usepackage{tcolorbox}
\newtcbox{\redbox}{on line,
  colframe=red,colback=white,
  boxrule=1pt,arc=1pt,boxsep=0pt,left=2pt,right=2pt,top=2pt,bottom=2pt}
\newcommand{\cmark}{\ding{51}}%
\usepackage{color-edits}
\addauthor{gn}{magenta}
\addauthor{zw}{orange}
\addauthor{df}{cyan}

\icmltitlerunning{\textsc{TroVE}: Inducing Verifiable and Efficient Toolboxes for Solving Programmatic Tasks}

\begin{document}

\twocolumn[
\icmltitle{\textsc{TroVE}: Inducing Verifiable and Efficient Toolboxes\\ for Solving Programmatic Tasks}



\icmlsetsymbol{equal}{*}

\begin{icmlauthorlist}
\icmlauthor{Zhiruo Wang}{cmu}
\icmlauthor{Graham Neubig}{cmu}
\icmlauthor{Daniel Fried}{cmu}
\end{icmlauthorlist}

\icmlaffiliation{cmu}{Language Technologies Institute, Carnegie Mellon University}

\icmlcorrespondingauthor{Zhiruo Wang}{zhiruow@cs.cmu.edu}

\icmlkeywords{Machine Learning, ICML}

\vskip 0.3in
]



\printAffiliationsAndNotice{}  % leave blank if no need to mention equal contribution

\begin{abstract}
Language models (LMs) can solve tasks such as answering questions about tables or images by writing programs. However, using primitive functions often leads to verbose and error-prone programs, and higher-level functions require expert design.
To enable better solutions without human labor, we ask code LMs to curate reusable high-level functions, and use them to write solutions.
We present \textsc{TroVE}, a \textit{\underline{tr}aining-free} method 
\underline{o}f inducing a \textit{\underline{ver}ifiable and \underline{e}fficient toolbox} of functions, by generating via using, growing, and periodically trimming the toolbox. 
On 11 datasets from math, table question answering, and image reasoning tasks, \textsc{TroVE} consistently yields \textit{simpler solutions} with \textit{higher accuracy} than baselines using \textsc{CodeLLaMa} and previous methods using \textsc{GPT}, while using $79$-$98\%$ \textit{smaller} toolboxes. \textsc{TroVE} further enables $31\%$ \textit{faster} and $13\%$ \textit{more accurate human verification} than baselines. With the same pipeline, it creates \textit{diverse functions} for varied tasks and datasets, providing insights into their individual characteristics. Code and data are available at \url{https://github.com/zorazrw/trove}.
\end{abstract}

\section{Introduction}

\begin{figure}[t]
\vspace{-1mm}
    \centering
    \includegraphics[width=0.49\textwidth]{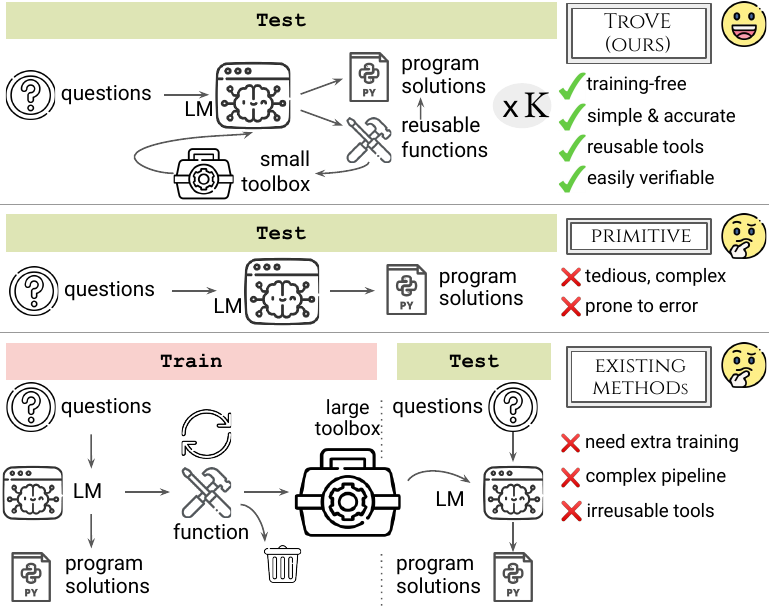}
    \vspace{-5mm}
    \caption{Our \textsc{TroVE} induces reusable functions to produce better program solutions than the \textsc{Primitive} setting, without training, supervision, or iterations required by \textsc{existing method}s.}
\vspace{-1mm}
\label{fig:teaser}
\end{figure}

\begin{figure*}[ht]
    \centering
    \includegraphics[width=\textwidth]{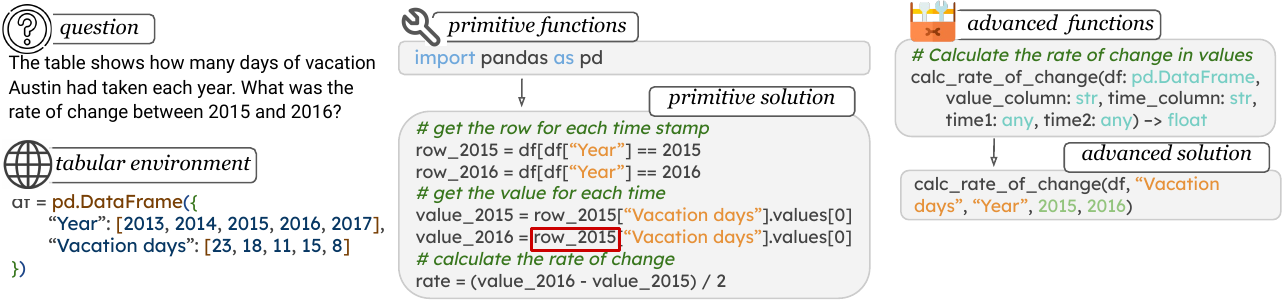}
    \vspace{-5mm}
    \caption{Function design affect solutions. Using primitive functions results in complex, error-prone solutions (middle), while using abstract functions leads to more concise and accurate solutions (right).}
\label{fig:problem-statement}
\end{figure*}

% solving problems via programs
Generating code from natural language commands has long been a method of choice for solving tasks such as question answering \citep{zettlemoyer2007online,liang2011learning} or agent navigation \citep{artzi2013weakly}. Recently, language models (LMs) have been used to write programs in general-purpose languages such as Python, further expanding code generation's applicability \citep{yin2017syntactic,li2022competition,cheng2023binding}. 
% primitive
These programs generally rely on multiple function calls to Python built-in functions or libraries such as \texttt{pandas}, as in the example in \autoref{fig:problem-statement}. However, in many cases relying on these primitive functions can lead to programs that are tedious, complex, and error-prone \citep{cai2023large,majumder2023clin}. These programs can also be difficult to verify, as the users may need to check every operation as well as their combinations and interactions across the entire program (e.g., the primitive solution in \autoref{fig:problem-statement}).

% advanced better than primitive
When human developers are faced with an analogous situation, they \textit{create application-specific functions}, i.e.~tools, composing primitive functions that are often used together.
For instance, in \autoref{fig:problem-statement} (right), the \texttt{calc\_rate\_of\_change} tool is easier to understand and less error-prone to use, hence enabling a more concise and accurate solution.

% model-induced functions
A few recent works have attempted to use LMs to \emph{automatically induce tools} in a similar way (\textsc{existing method}s in \autoref{fig:teaser}).
However, existing methods tend to either induce large and ponderous toolboxes and/or have added complexity and data requirements.
For instance, \citet{qian2023creator} propose CREATOR to disentangle planning (tool making) from execution, but at the cost of producing hundreds or more tools that are challenging for models to reuse or humans to verify. 
Further, compared to the standard setting (\textsc{primitive} in \autoref{fig:teaser}) that only needs test data to produce their solutions, \citet{wang2023voyager} propose life-long learning via an automatic curriculum, equipped with iterative self-verification. \citet{cai2023large,yuan2023craft} require additional training and validation datasets to create tools ahead to be used by solutions, plus auxiliary modules such as self-verification or toolset deduplication.

% intro of our method
% We show that it is possible to induce a sufficiently small number of reusable functions that lead to more accurate solutions online at test time.
In this paper, we propose \textsc{TroVE}, a \underline{tr}aining-free method \underline{o}f inducing a \underline{v}erifiable and \underline{e}fficient function toolbox (\S\ref{sec:4:method}), and using these functions to write solutions.
\textsc{TroVE} features three major components: using and growing a toolbox maintained over time, execution agreement-based selection, and periodic toolbox trimming.
Notably, our method \textit{requires zero additional training or supervision}, and selects programs only by their inter-\textit{execution agreement}. Given a stream of questions, \textsc{TroVE} produces their solutions, along with inducing a handy toolbox to solve the questions.

% experiment & results
We experiment on 11 datasets from three real-world tasks (\S\ref{sec:5:dataset}): (1) mathematical problems with the MATH dataset, (2) table question answering on TabMWP, WTQ, and HiTab, and (3) compositional visual reasoning with GQA.
Compared to baselines using \textsc{CodeLLaMa} as well as previous state-of-the-art methods using \textsc{GPT}, our \textsc{TroVE} consistently produces solutions with higher accuracy and reduced complexity, while maintaining a significantly smaller, efficient function library (\S\ref{sec:6:experiment}).
We further show that via human study (\S\ref{sec:7:human-verify}), verifying solutions generated by \textsc{TroVE} is $31\%$ faster and $13\%$ more accurate, than solutions generated by baseline methods.
% Moreover, while \textsc{TroVE} is generically applicable to all datasets and tasks involved, it ends up creating appropriate functions varying in sources \gncomment{I didn't understand ``sources''} and functionalities \gncomment{``functionality'' is typically just singular.} for different domains (\S\ref{sec:8:special-function}).
% Lastly, we perform in-depth analyses (\S\ref{sec:9:ablation-study}) and show that \textsc{TroVE} (1) is robust to example ordering and (2) benefits from tunable toolbox trimming.

\section{Problem Statement \& Baseline Methods}
\label{sec:3:problem-statement}

We formally define the task of problem solving via programs (\S\ref{sub:3.1:problem-solving}) and introduce corresponding baseline methods (\S\ref{sub:3.2:baseline-methods}).

% ##################### %
\subsection{Problem Solving via Programs}
\label{sub:3.1:problem-solving}
We focus on problems that are describable in natural language (NL) and solvable using programs. Concretely, given an example $x$, i.e., an NL query $q$ grounded on an environment $e$, we ask a language model $LM$ to write a programmatic solution $s$ by composing multiple functions $F = \{f_1, \cdots, f_n\}$. This process is denoted as $P_{LM}(q, e, f)$. $f$ and hence $s$ can be executed on $e$ to obtain the final answer.
We use Python as the programming language for our experiments, since it is general-purpose thus allowing flexible functions to be created for most tasks.
Each solution $s$ is a Python program, and each function $f$ is a Python function.

It is crucial to note that the difficulty of solution generation is greatly affected by the usability of the functions in the set $F$.
Relatively speaking, we can categorize functions into two types: (i) \textit{primitive functions}, which only support basic operations such as Python standard libraries, and (ii) \textit{composed functions}, which perform more complex operations by composing multiple basic operations.

% ######### %
\paragraph{Primitive Functions} 
Primitive functions are atomic, low-level operations on the task environment, such as subtraction \colorbox{gray!30}{\texttt{-}} and division \colorbox{gray!30}{\texttt{/}} in the \textit{primitive solution} in \autoref{fig:problem-statement} (middle). They are often easy to obtain without expert knowledge about the application domain.
Yet often, to solve a question as in \autoref{fig:problem-statement} (left), it requires complex compositions of numerous functions and hence becomes extremely error-prone. In this example, due only to a tiny \redbox{mistake}, which calls the wrong \texttt{row\_2015} instead of \texttt{row\_2016}, the output goes wrong despite all the other steps being correct.

% ######### %
\paragraph{Composed Functions}
Composed functions, such as the \texttt{calc\_rate\_of\_change} in \autoref{fig:problem-statement} (right), combine various primitive functions.
Conceptually, they are \textit{easier to use}, as they align better with the actions asked in the question.
In this example, it is easier to associate ``What is the rate of change ...'' with a function named \texttt{calc\_rate\_of\_change}, compared to certain compositions of data slicing \texttt{df[$\cdot$]}, check equality \texttt{==}, get value \texttt{cell.values[index]}, subtraction \texttt{-}, and division \texttt{/}.
Practically, composed functions are \textit{less error-prone}, by only showing an API interface and abstracting intricate details inside. 

Composed functions are often crafted by human experts, by recognizing and generalizing shared functionality. However, this process is costly and hardly scalable to new domains. Therefore, it is crucial to create such functions automatically, to enhance problem solving while saving human labor.

% ##################### %
\subsection{Baseline Methods}
\label{sub:3.2:baseline-methods}

We introduce two baselines that generate programs using primitive and composed functions.
All methods, including our main method later in \S\ref{sec:4:method}, operate solely by prompting an LM, and do not update the parameters of the LM itself.

% ######### %
\paragraph{Using Primitive Functions}
Our first baseline, \textsc{Primitive}, asks models to generate programs using primitive functions, which is the de facto approach for program-aided problem solving without tool induction \citep{cheng2023binding,gao2023pal}.

\begin{figure}[ht]
    \centering
    \includegraphics[width=0.39\textwidth]{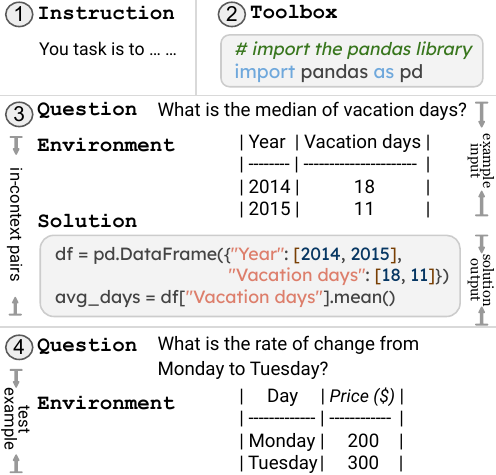}
    \vspace{-2mm}
    \caption{Example input prompt on the tabular environment. We textualize tables by markdown format in the prompt, and ask LMs to parse tables into \texttt{pandas} DataFrame in program solutions.}
\vspace{-1mm}
\label{fig:prompt-input}
\end{figure}

As exemplified in \autoref{fig:prompt-input}, our prompt inputs consist of four components: (1) an NL instruction specifying the task, (2) the function signature and textual docstring of primitive functions,\footnote{We do not show the built-in functions since LMs have already been trained on them extensively. We only have 1-2 primitives in addition (\S\ref{sub:5.3:data-query}, \S\ref{sub:5.4:image-reasoning}), so they can easily fit into the prompt.}
(3) $c$ (example, solution) pairs to demonstrate the usage of primitive functions, and lastly (4) the query and environment ($q, e$) of the current testing example. We collect the code snippets from model responses as $s$ and execute on $e$ to obtain the final result and evaluate it.
Please find more detailed examples in \S\ref{app:a:prompt-example}.

% ######### %
\paragraph{Abstracting Functions Example Wise}

Our second baseline \textsc{Instance} asks models to create tools for each example, and use them in the solution for that example.
\citet{qian2023creator} found this two-step process helpful for model reasoning by disentangling tool abstraction (planning) from example-wise decision (execution). 

\begin{figure}[ht]
    \centering
    \includegraphics[width=0.39\textwidth]{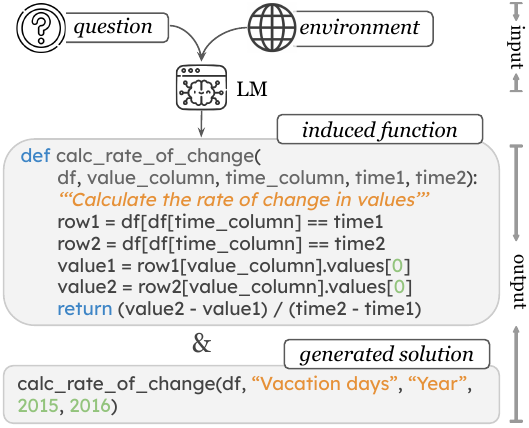}
    \vspace{-2mm}
    \caption{An illustration of a model inducing a composed function and using it to generate the solution at the same time.}
% \vspace{-1mm}
\label{fig:dual-generation}
\end{figure}

In preliminary studies, we compared generating functions and solutions in two sequential responses or prompting the model to generate both in one response.
They performed comparably, so we adopted the latter approach since it is simpler.
Concretely, given a set of primitive functions $P$, for each example ($q, e$), the model needs to generate the functions $F$ by composing operations in $P$, as well as the solution $s$ that uses $F$. Grounding on \autoref{fig:dual-generation}, $F$ is the \textit{induced function} snippet, and $s$ is the \textit{generated solution}.

We include the same four prompt components used in \textsc{Primitive}, but alter the instruction and example outputs according to the $F\&s$ generation format. We query each example independently to allow example-wise function induction. 
However, it is not possible to share these functions across examples, even if they have similar functions. 
% In \S\ref{sec:4:method}, we present \textsc{TroVE} that maintains a shared toolbox over time.
\section{Inducing Reusable Functions On-the-fly}
\label{sec:4:method}

Now, we present our main method \textsc{TroVE} that: uses and grows a toolbox iteratively over time (\S\ref{sub:4.1:multi-channel}), selects optimal outputs based on execution agreement (\S\ref{sub:4.2:agreement-selection}), and periodically trims low-utility functions from the toolbox (\S\ref{sub:4.3:forgetting-mechanism}).

% ##################### %
\subsection{Using and Growing the Toolbox}
\label{sub:4.1:multi-channel}

\begin{figure}[ht]
    \centering
    \includegraphics[width=0.41\textwidth]{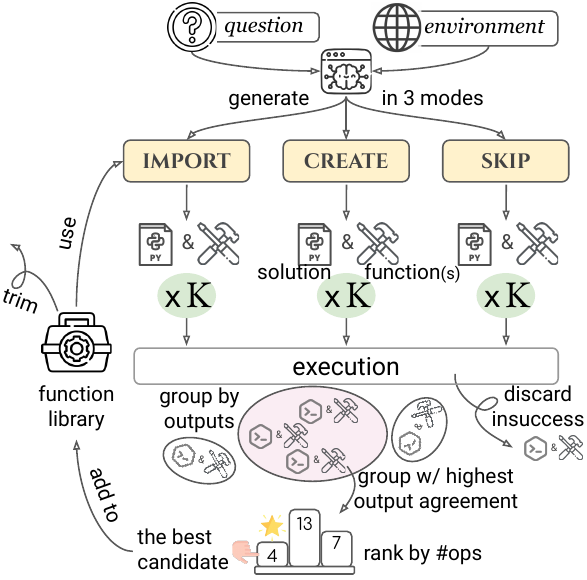}
    \vspace{-2mm}
    \caption{\textsc{TroVE} illustration. Top: generate solutions while using and growing the toolbox. Bottom: select the best response by execution agreement. Left: periodically trim low-utility functions.}
\vspace{-1mm}
\label{fig:online-pipeline}
\end{figure}

To learn functions that can be reused across examples, we maintain a shared function library $F$ over time. 
To keep our method running in linear time, we process all examples \textit{online} in a streaming fashion.
We start with $F = \emptyset$ and gradually add or remove functions from it.
\autoref{fig:online-pipeline} illustrates the processing of example $x^t$ at time $t$.

First, we define 3 modes with which LMs can interact with the current toolbox $F^t$. 
\circled{1} \colorbox{amber!27}{\textsc{IMPORT}}: In this mode, the LM is instructed to import defined functions from the toolbox and apply them to solve $x^t$.
\circled{2} \colorbox{amber!27}{\textsc{CREATE}}: In this mode, the LM is instructed to create a new function $f^t_{new}$ and add it to the toolbox.
\circled{3} \colorbox{amber!27}{\textsc{SKIP}}:
In this mode, the LM is instructed to only use primitive functions just as in the \textsc{Primitive} setting.

For each example, for each of the three modes we sample from the LM to generate $K$ responses, for a total of $3K$ responses per example.\footnote{In preliminary studies, we tried to let models choose one mode and only generate in that mode, but results degraded significantly.}
Each response contains a solution $s$ and function $f$ as shown in \autoref{fig:dual-generation}, except for the \textsc{SKIP} mode that only contains $s$.
We denote these candidate responses as $(f_m^{ti}, s_m^{ti})$, where $m \in \{\textsc{import}, \textsc{create}, \textsc{skip} \}$ and $i \in \{1, \cdots, K \}$.

% ##################### %
\subsection{Agreement-Based Selection}
\label{sub:4.2:agreement-selection}
From the $3K$ sampled $(f, s)$ pairs, we select one to use via self-consistency \citep{shi-etal-2022-mbr-exec,li-2022-alpha-code,wang2023selfconsistency} and solution complexity.
We first execute all solutions and remove those that cannot execute successfully. Next, we select an answer based on the consistency of the execution outputs, keeping the solutions that produce the most frequently occurring answer. Next, if there are multiple solutions with the same frequency, we rank solutions by the number of operations they require, and pick the one with the least operations (preferring simple solutions). Finally, if multiple candidates remain, we break the tie by arbitrarily choosing the one that appears first in the model prediction list.
We add the function in this best response to the toolbox, and adopt its solution as the solution for the current example. 
% In short, our selection heuristic prefers candidates with high agreement and simpler solutions.

% ##################### %
\subsection{Periodic Toolbox Trimming}
\label{sub:4.3:forgetting-mechanism}

Not all the functions induced by models are highly reusable. Hence, we also propose to periodically trim the toolbox to effectively remove low-utility functions.

Periodically during testing, we remove functions that have been used less than $\lambda$ times. By observing that function-usage frequency has a logarithmic relation with data size, we set $\lambda = C \times \log_{10}(n)$, where $C = \frac{1}{2}$, $n$ is the number of examples processed so far.\footnote{To ensure all examples can be solved by functions available in the library, we update the solutions for examples previously using trimmed functions ($<5\%$ of the dataset), by re-generating solutions under \textsc{IMPORT} \& \textsc{SKIP} modes.}
% To ensure all examples can be solved by functions available in the library, we update the solutions for examples previously using trimmed functions, by re-generating solutions under \textsc{IMPORT} \& \textsc{SKIP} modes.\footnote{These examples usually take less than 5\% of the dataset.} We evaluate solutions after this update.

% Our goal is to prioritize generic functions that can solve more examples. While low-utility functions serve more as a distraction during function selection, periodic trimming helps model to focus on potentially more helpful functions.

\section{Testbeds: Program-Solvable Tasks}
\label{sec:5:dataset}

We now introduce the three programmatic tasks for experiments: math problems, table question answering, and visual reasoning. 
We first state the default primitive functions (\S\ref{sub:5.1:common-primitive}), then describe the specialized functions for each task.

% ##################### %
\subsection{The Default Primitive Functions}
\label{sub:5.1:common-primitive}
For all experiments, we instruct models to generate solutions as Python programs, so that default primitive functions are built-in Python functions.\footnote{\url{https://docs.python.org/3/library/functions.html}}
We use these default primitives for MATH (\S\ref{sub:5.2:math-problems}), and add other data-related functions for TableQA (\S\ref{sub:5.3:data-query}) and VisualQA (\S\ref{sub:5.4:image-reasoning}).

% ##################### %
\subsection{Math Problems}
\label{sub:5.2:math-problems}
To test model abilities in solving math problems, we use the MATH \citep{hendrycks2021measuring} dataset that covers questions from seven subjects: algebra, counting and probability, geometry, intermediate algebra, number theory, prealgebra, and precalculus. \autoref{tab:dataset-stats} lists the number of test examples, since our methods only need test data. 
We only use the default primitives, i.e., built-in Python functions.

\begin{table}[ht]
% \vspace{2mm}
\small 
\centering 
\resizebox{0.5\textwidth}{!}{
    \begin{tabular}{llrl}
    \toprule
    \textbf{Task} & \textbf{Dataset} & \textbf{Size} & \textbf{Primitive Functions} \\
    \midrule 
    \multirow{7}{*}{\textsc{MATH}} & {algebra} & {881} & \multirow{7}{*}{\texttt{built-in functions}} \\
    {} & {count \& prob.} & {291} & {} \\
    {} & {geometry} & {237} & {} \\
    {} & {inter. algebra} & {503} & {} \\
    {} & {number theory} & {497} & {} \\
    {} & {prealgebra} & {636} & {} \\
    {} & {precalculus} & {156} & {} \\
    \midrule
    \multirow{4}{*}{\textsc{TableQA}} & {TabMWP} & {5,376} & {+ \texttt{pandas}} \\
    \cmidrule{3-4}
    {} & {WTQ} & {4,344} & {+ \texttt{pandas}} \\
    \cmidrule{3-4}
    {} & \multirow{2}{*}{HiTab} & \multirow{2}{*}{1,574} & {+ \texttt{pandas}} \\
    {} & {} & {} & {+ \texttt{parse\_table}} \\
    \midrule
    \multirow{4}{*}{\textsc{VisualQA}} & \multirow{4}{*}{GQA} & \multirow{4}{*}{12,578} & {+ \texttt{PIL.Image}} \\
    {} & {} & {} & {+ \texttt{locate\_objects}} \\
    {} & {} & {} & {+ \texttt{visual\_qa}} \\
    {} & {} & {} & {+ \texttt{crop\_region}} \\
    \bottomrule
    \end{tabular}
}
\vspace{-2mm}
\caption{Statistics and primitives for three tasks.}
\vspace{-2mm}
\label{tab:dataset-stats}
\end{table}

\begin{table*}[ht]
\small 
\centering 
\resizebox{\textwidth}{!}{
    \begin{tabular}{lc|rrrrrrr|rrr|r}
    \toprule
    \multirow{2}{*}{\textbf{Method}} & \multirow{2}{*}{\textbf{Metric}} & \multicolumn{7}{c|}{\textsc{MATH}} & \multicolumn{3}{c|}{\textsc{TableQA}} & \multicolumn{1}{c}{\textsc{Visual}} \\
    {} & {} & {alg} & {count} & {geo} & {inte} & {num} & {prealg} & {precal} & \multicolumn{1}{c}{TabMWP} & \multicolumn{1}{c}{WTQ} & \multicolumn{1}{c|}{HiTab} & \multicolumn{1}{c}{GQA} \\
    \midrule
    \multirow{3}{*}{\textsc{Primitive}} & {acc $~~\uparrow$} & {0.15} & {0.14} & {0.06} & {0.05} & {0.16} & {0.21} & {0.10} & {0.43} & {0.20} & {0.09} & {0.37} \\
    {} & {\# ops $\downarrow$} & \textbf{15.4} & {10.9} & \textbf{15.1} & \textbf{17.0} & {12.3} & {12.1} & {20.8} & {17.4} & {24.3} & {16.5} & {24.8} \\
    {} & {\# lib $~\downarrow$} & \multicolumn{7}{c|}{---} & \multicolumn{3}{c|}{---} & \multicolumn{1}{c}{---} \\
    \midrule
    \multirow{3}{*}{\textsc{instance}} & {acc $~~\uparrow$} & {0.22} & {0.23} & {0.07} & {0.06} & {0.23} & {0.26} & \textbf{0.17} & {0.36} & {0.17} & {0.12} & {0.16} \\
    {} & {\# ops $\downarrow$} & {18.4} & {10.2} & {26.8} & {28.2} & {14.3} & \textbf{10.6} & {26.9} & \textbf{8.3} & \textbf{8.4} & {14.1} & \textbf{18.8} \\
    {} & {\# lib $~\downarrow$} & {39} & {7} & {36} & {82} & {5} & {16} & {36} & {3,175} & {537} & {31} & {395} \\
    \midrule
    \multirow{3}{*}{\textsc{TroVE}} & {acc $~~\uparrow$} & \textbf{0.25} & \textbf{0.26} & \textbf{0.08} & \textbf{0.11} & \textbf{0.25} & \textbf{0.29} & \textbf{0.17} & \textbf{0.47} & \textbf{0.21} & \textbf{0.18} & \textbf{0.44} \\
    {} & {\# ops $\downarrow$} & {18.8} & \textbf{10.0} & {25.4} & {23.9} & \textbf{11.2} & {11.7} & \textbf{19.6} & {10.9} & {9.2} & \textbf{}{9.3} & {20.3} \\
    {} & {\# lib $~\downarrow$} & {10} & {1} & {7} & {8} & {8} & {4} & {7} & {10} & {11} & {5} & {7} \\
    \bottomrule
    \end{tabular}
}
\vspace{-2mm}
\caption{\textsc{CodeLLaMa-7b-Instruct} results on \textsc{MATH}, \textsc{TableQA}, and \textsc{Visual} tasks.}
\label{tab:main-results}
\end{table*}

% ##################### %
\subsection{Table Question Answering}
\label{sub:5.3:data-query}
We adopt three table question answering datasets: TabMWP \citep{lu2023dynamic}, WTQ \citep{pasupat-liang-2015-compositional}, and Hitab \citep{cheng-etal-2022-hitab}. They cover a diverse range of question types and table structures.
We represent tables as \texttt{pandas} DataFrame objects since this is a standard table library to use with Python; we accordingly add the \texttt{pandas} library into the set of primitive functions for the table QA task.

\paragraph{TabMWP}
The TabMWP dataset \citep{lu2023dynamic} includes math word problems on relational tables. Questions in TabMWP are relatively simple, such as performing numerical calculations (e.g., ``What is the mean of the numbers?'') and argument selection (``Who has the most OBJECT?''). 
We directly input the serialized tables in markdown format in model prompts, because they are relatively small.

\paragraph{WTQ}
The WikiTableQuestions (WTQ) dataset \citep{pasupat-liang-2015-compositional} contains questions about semi-structured Wikipedia tables, which feature un-normalized cell values, thus requires string processing (e.g., parse the number from ``\$ 100.00'') and external knowledge retrieval (e.g., find the country name of ``Franco Pellizotti (ITA)'') operations.

Because WTQ tables can be too long to input limits, we put DataFrame previews in the prompt, and instruct models to use \texttt{pandas.read\_table} to load tables from CSV files.

\paragraph{HiTab}
The Hitab dataset \citep{cheng-etal-2022-hitab} contains questions about hierarchical matrix tables. HiTab tables have more complex structures and require special operations such as multi-hop selection along a bi-dimensional header hierarchy. 
We similarly load HiTab tables from the source JSON files using the \texttt{parse\_table} function used by \citet{cao2023apiassisted}, and add it as the primitive functions.

% More details of table processing are in \S\ref{app:c:data-processing}.

% ##################### %
\subsection{Visual Reasoning}
\label{sub:5.4:image-reasoning}
We use the GQA dataset \citep{hudson2019gqa} that contains real-world images and compositional questions about them.
However, the skills required to solve GQA questions are more advanced than the previous two tasks. Although image processing libraries such as \texttt{PIL} or \texttt{cv2} are available, our preliminary experiments show that using these libraries alone is extremely hard for models to generate viable solutions (only achieving 1\% accuracy), not to mention further inducing advanced functions. 

Therefore, we adopt three main primitive actions from VisProg \citep{gupta2022visual} of two types:
(1) neural modules: \texttt{visual\_qa}, \texttt{locate\_objects}; and
(2) processing modules: \texttt{crop\_region}; along with the image loading module \texttt{PIL.Image}.
To reduce the extent of expert engineering, we did not adopt the other six actions used in VisProg, since they may be tailored to GQA queries or crafted shortcuts to assist models (e.g., \texttt{check\_exists}). Removing these actions slightly degrades model performance.

\section{Experiments}
\label{sec:6:experiment}

We introduce the experiment setup (\S\ref{sub:6.1:setup}) and evaluation metrics (\S\ref{sub:6.2:eval}), then report and analyze the results (\S\ref{sub:6.3:results}).

% ##################### %
\subsection{Experimental Setup}
\label{sub:6.1:setup}
We compare our method \textsc{TroVE} (\S\ref{sec:4:method}) to the two baselines \textsc{Primitive} and \textsc{Instance} (\S\ref{sub:3.2:baseline-methods}).
We mainly use \textsc{CodeLLaMa2-7b-Instruct} for experiments,\footnote{We are the first to show that open-source LMs can make tools.} but also use \textsc{GPT-4} to fairly compare with existing SOTA methods.
We include $c = 2$ examples in prompts, sampled $K = 5$ responses in each mode, and trim the toolbox every 200 steps.
By default, we set the decoding temperature to $0.6$ and use top-p $0.95$. We limit the model to generate at most 512 tokens to prevent excessive hallucination and save computational cost. 
To accommodate for randomness in the sampling result, we run each experiment five times and report the best-performing run. 
For all methods, we evaluate the results on test examples.

% ##################### %
\subsection{Evaluation Metrics}
\label{sub:6.2:eval}
We propose three metrics to comprehensively evaluate generated solutions and induced functions.

\paragraph{Answer Correctness} (acc $\uparrow$)
The most practically important aspect of solutions is correctness. We measure if the execution outcome of the solution program exactly matches the ground-truth answer(s). 

\paragraph{Solution Complexity} (\# ops $\downarrow$)
We also measure the program complexity by counting the number of function calls involved. Solutions with fewer functions are easier and quicker to understand and verify. %, and cost less at inference. %The lower the complexity, the better the solutions are.

\paragraph{Library Size} (\# lib $\downarrow$)
It is important to control the library size and encourage function sharing across examples. 
Compared to multiple tools performing similar operations, fewer tools with distinct functionaly enable easier tool selection during solution generation.
Since the number of primitive functions is always the same on a given dataset, we only report the number of additionally induced functions.

% Refer to more implementation details in \S\ref{app:b:metric-detail}.

% ##################### %
\subsection{Model Performance}
\label{sub:6.3:results}

\autoref{tab:main-results} shows the results on all datasets. %\autoref{fig:qa-results} visualizes the \textsc{TableQA} and \textsc{Visual} results in \autoref{tab:main-results}.
\textsc{TroVE} produces the most accurate solutions with generally lower complexity, while maintaining a small, efficient function library.

\paragraph{Math Problems}
Compared to \textsc{Primitive}, \textsc{TroVE} substantially improves answer correctness by 30--120\% across 7 datasets.
Comparing to \textsc{Instance}, \textsc{TroVE} yields 8.7--83.3\% higher correctness using 60.0--90.2\% fewer tools.

\paragraph{Table Question Answering}
% As illustrated in \autoref{fig:qa-results}, \textsc{TroVE} also produces the most accurate solutions (the rightmost) with the least complexity (the bottom), with an efficient library (small circle).
% \textsc{TroVE} also produces the most accurate solutions with the least complexity, with an efficient library.
Notably, \textsc{instance} yields lower correctness than \textsc{Primitive} on most datasets, while generating hundreds even thousands of functions. We conjecture the reason to be increased task difficulty and dataset size (than \textsc{MATH}), driving up the number of functions and confusing the model with too many low-utility options. While \textsc{TroVE}, by reusing and trimming functions, alleviates this distraction and gives the best results.

\paragraph{Visual Question Answering}
\textsc{TroVE} still scores the best, but \textsc{instance} performs substantially worse than other methods. with a correctness drop of $0.21$ compared to \textsc{Primitive}. Similarly to \textsc{TableQA}, a larger number of functions (i.e., 395) are created, which greatly challenges the solution generation process. Based on our result analysis, we conjecture that it is difficult to create many valid reusable functions for GQA, hence most induced functions are invalid and impair solution generation.

\subsection{Comparing with Other Tool-Making Methods}
In addition, we compare \textsc{TroVE} with three existing methods that perform tool making, namely LATM \citep{cai2023large}, CREATOR \citep{qian2023creator}, and CRAFT \citep{yuan2023craft}.
Notably, all three methods include extra modules or supervision not required by our method, and were only demonstrated effective on \textit{closed-source} GPT models.
Specifically, LATM requires an extra training set to induce tools in advance, and a validation set to verify the tools. 
CREATOR runs multiple iterations of decision, execution, and rectification.
CRAFT requires extra training data and tool retrieval modules, and necessitates \textsc{GPT} models.

To make a fair comparison, we also use \textsc{GPT-4} with our \textsc{TroVE} approach, and test on three datasets --- algebra from \textsc{MATH}, TabMWP from \textsc{TableQA}, and GQA from \textsc{VisualQA} task --- that overlap with these works. Due to resource limitations, we do not experiment on all 11 datasets. We believe the result differences in these 2 datasets are representative to demonstrate the superiority of our method.

\begin{table}[ht]
\vspace{1mm}
\small 
\centering 
\resizebox{0.49\textwidth}{!}{
    \begin{tabular}{l|cc|cc|cc}
    \toprule
    \multirow{2}{*}{\textbf{Method}} & \multicolumn{2}{c|}{\textbf{MATH$_{algebra}$}} & \multicolumn{2}{c|}{\textbf{TabMWP}} & \multicolumn{2}{c}{\textbf{GQA}} \\
    {} & {acc $\uparrow$} & {\# lib $\downarrow$} & {acc $\uparrow$} & {\# lib $\downarrow$} & {acc $\uparrow$} & {\# lib $\downarrow$} \\
    \midrule
    \multicolumn{4}{l}{\textit{w/ additional supervision}} \\
    {LATM} & {0.30} & {-} & {0.09} & {-} & {0.29}  & {-} \\
    {CRAFT} & {0.68} & {282} & {0.88} & {181} & \textbf{0.45} & {525} \\
    \midrule
    \multicolumn{4}{l}{\textit{w/ additional rectification \& iteration}} \\
    {Creator} & {0.65} & {875} & {0.81} & {4,595} & {0.34} & {-} \\
    \midrule
    \multicolumn{4}{l}{\textit{w/o supervision, rectification, or iteration}} \\
    \textsc{TroVE} & \textbf{0.72} & \textbf{16} & \textbf{0.92} & \textbf{38} & {0.44} & \textbf{8} \\
    \bottomrule
    \end{tabular}
}
\vspace{-2mm}
\caption{Comparing with existing methods using \textsc{GPT-4}. We adopt the baseline results as reported in \citet{yuan2023craft}. We do not report the \textit{complexity} metric since none of these methods report it (our results in \autoref{tab:main-results}).}
\vspace{2mm}
\label{tab:tool-baseline}
\end{table}

In \autoref{tab:tool-baseline}, \textsc{TroVE} outperforms these existing state-of-the-art methods on all datasets and most evaluation aspects, while having a much simpler pipeline. \textsc{TroVE} not only works with closed-source \textsc{GPT}s, but also open-source \textsc{CodeLLaMa} (\autoref{tab:main-results}), with which CRAFT reported near-random performance using their method.

% ##################### %
\paragraph{Training Advantage of Seen Primitive Functions}

While \textsc{GPT-4} outperforms \textsc{CodeLLaMa} on most tasks, it is impressive to see that they perform comparably on the GQA task, as shown in \autoref{tab:train-primitive}.
We conjecture that this difference between GQA and other tasks comes from the advantage of models using corresponding primitive functions. For example, \texttt{pandas}, as a primitive in \textsc{TableQA}, may appear frequently in the training data, so models may be more proficient in or inclined to write solutions with this library. 

In contrast, models may have never seen any data using GQA primitives since no large-scale data are annotated with these hand-crafted functions. The difficulty of models learning these primitives in context is similar to that of learning the induced functions.
While GQA reduces GPT-4's advantage in using primitives, it is intriguing to observe that 7B \textsc{CodeLlama2} performs on par with \textsc{GPT-4} by making and (re-)using tools.

\begin{table}[ht]
\vspace{-1mm}
\small 
\centering 
\resizebox{0.45\textwidth}{!}{
    \begin{tabular}{ll|ccc}
    \toprule
    \multirow{2}{*}{\textbf{Model}} & \multirow{2}{*}{\textbf{Method}} & \multicolumn{3}{c}{\textbf{Evaluation Metrics}} \\
    {} & {} & {acc $\uparrow$} & {\# ops $\downarrow$} & {\# lib $\downarrow$} \\
    \midrule
    \multirow{2}{*}{\textsc{CodeLLaMa}} & \textsc{Primitive} & {0.37} & {24.6} & {-} \\
    {} & \textsc{TroVE} & {0.44} & {20.3} & {7} \\
    \midrule
    \multirow{2}{*}{\textsc{GPT-4}} & \textsc{Primitive} & {0.40} & {27.4} & {-} \\
    {} & \textsc{TroVE} & {0.44} & {20.2} & {8} \\
    \bottomrule
    \end{tabular}
}
\vspace{-1mm}
\caption{7B \textsc{CodeLlama2} and \textsc{GPT-4} perform comparably on the GQA task without training advantage.}
\vspace{-1mm}
\label{tab:train-primitive}
\end{table}

\section{Efficient Verification by Humans}
\label{sec:7:human-verify}
Model-produced solutions may not be reliable, so we investigate if \textsc{TroVE} facilitates more efficient solution verification by humans.
To test this hypothesis, we randomly selected 100 examples and asked 6 human evaluators to verify the correctness of solutions generated by \textsc{Primitive}, \textsc{Instance}, and \textsc{TroVE} on the WTQ dataset.\footnote{We chose WTQ from the \textsc{TableQA} task, because \textsc{TableQA} functions are more complex than those in other tasks, and WTQ is the fairest representative of the task with mid-level difficulty.}

We evaluate human performance from two aspects. (1) Detection accuracy: if they can accurately predict whether or not the solution is correct; the higher the better, and (2) Time used: how many seconds they need to verify an average example; the lower the better.

\autoref{tab:human-verify} shows the results.
For \textit{accuracy}, using tools in solutions (\textsc{Instance} and \textsc{Trove}) improves detection accuracy by 13.0--14.3\%, compared to using \textsc{Primitive} functions only.
For the \textit{time used}, our method reduces the average time by $31.4 \%$ compared to \textsc{primitive}, and $43.0 \%$ compared to \textsc{instance}. However, using irreusable tools (i.e., the \textsc{instance} setting) actually increases the time by $20.4 \%$.
Overall, \textsc{TroVE} substantially speeds up the verification process, while achieving similar detection accuracy.
See \S\ref{app:c:human-study} for the test results of individual participants.

\begin{table}[ht]
\small 
\centering 
\resizebox{0.40\textwidth}{!}{
    \begin{tabular}{l|cc|cc}
    \toprule
    \multirow{2}{*}{\textbf{Method}} & \multicolumn{2}{c|}{\textbf{Accuracy $\uparrow$}} & \multicolumn{2}{c}{\textbf{Time (s) $\downarrow$}} \\
    {} & {avg} & {std} & {avg} & {std} \\
    \midrule
    \textsc{Primitive} & {0.77} & {0.109} & {25.5} & {$~~$6.671} \\
    \textsc{Instance} & {0.88} & {0.024} & {30.7} & {12.750} \\
    \textsc{TroVE} & {0.87} & {0.057} & {17.5} & {$~~$4.855} \\
    \bottomrule
    \end{tabular}
}
\vspace{-1mm}
\caption{Human accuracy and time in verifying model-produced solutions with three methods experimented.}
\vspace{-1mm}
\label{tab:human-verify}
\end{table}

% ##################### %
\section{Inducing Specialized Functions}
\label{sec:8:special-function}
\textsc{TroVE} demonstrates its generality on multiple tasks and datasets. In this section, we further show that \textsc{TroVE} can produce specialized functions that (1) differ in forms across tasks, and (2) differ in functions across datasets.
We use the \textsc{CodeLLaMa2-7B} results as an example.

\paragraph{Different Function Forms Across Tasks}
We compare the three tasks and list a few exemplar functions in \autoref{tab:task-tools}. 
For \textsc{MATH}, the model often imports external libraries (\texttt{sympy}) to enable using advanced functions, or creates functions targeting certain problems (\texttt{calculate\_remainder}). \textsc{TableQA} tasks induce more complex functions comprising many primitive functions (e.g., \texttt{get\_match\_after\_condition}). \textsc{VisualQA} functions involve fewer primitives, for example, \texttt{get\_image\_region} is a chain of two primitives: \texttt{locate\_objects} and \texttt{crop\_region}.

\begin{table}[ht]
\vspace{1mm}
\small 
\centering 
\resizebox{0.49\textwidth}{!}{
    \begin{tabular}{l|l}
    \toprule
    \multicolumn{1}{c|}{\textbf{Task}} & \multicolumn{1}{c}{\textbf{Example Functions}} \\
    \midrule 
    \textsc{MATH} & {\adjustbox{valign=c}{\includegraphics[width=0.42\textwidth]{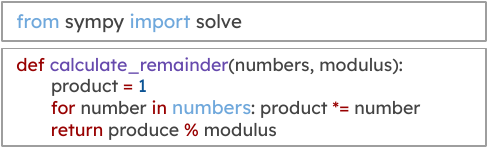}}} \\
    \midrule
    \textsc{TableQA} & {\adjustbox{valign=c}{\includegraphics[width=0.42\textwidth]{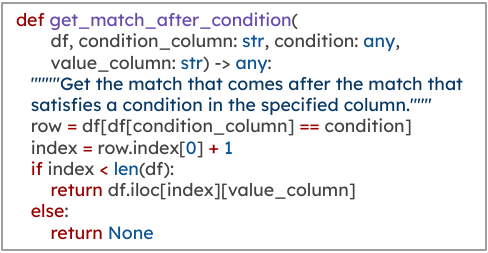}}} \\
    \midrule
    \textsc{VisualQA} & {\adjustbox{valign=c}{\includegraphics[width=0.42\textwidth]{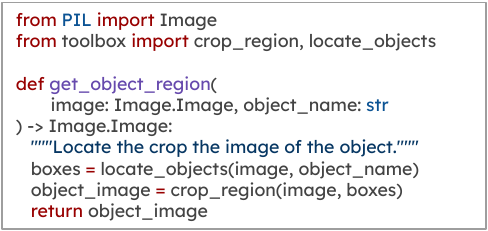}}} \\
    \bottomrule
    \end{tabular}
}
\vspace{-1mm}
\caption{Example functions induced by \textsc{TroVE} on three tasks.}
% \vspace{-1mm}
\label{tab:task-tools}
\end{table}

% ####### %
\paragraph{Varied Functionalities Across Datasets}
For \textsc{MATH} and \textsc{TableQA} that have multiple datasets, we further analyze the variance in functions between the datasets.

\begin{figure}[ht]
\vspace{-2mm}
    \centering
    \includegraphics[width=0.42\textwidth]{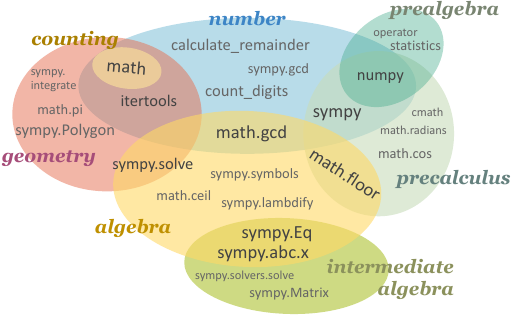}
    \vspace{-2mm}
    \caption{Illustration of \textsc{MATH} libraries for seven subjects.}
\vspace{-1mm}
\label{fig:math-toolbox}
\end{figure}

Among \textsc{MATH} datasets, as shown in \autoref{fig:math-toolbox}, core functions such as \texttt{math} and \texttt{sympy} overlap. 
Meanwhile, some functions are particularly useful for certain questions, such as the self-defined \texttt{calculuate\_remainder} for \textit{number theory} questions, and \texttt{sympy.Polygen} for \textit{geometry} questions.

\begin{figure}[ht]
\vspace{-1mm}
    \centering
    \includegraphics[width=0.40\textwidth]{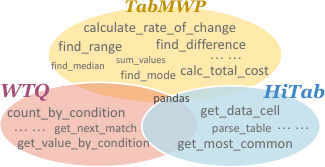}
    \vspace{-2mm}
    \caption{Illustration of three \textsc{TableQA} function libraries.}
\label{fig:table-toolbox}
\end{figure}

\autoref{fig:table-toolbox} shows some functions in three \textsc{TableQA} datasets. Due to the greater variance in question types and table structures, most functions differ except for the basic \texttt{pandas}.

Overall, \textsc{TroVE} can effectively propose both functions that are (1) generic to the task, and (2) specific to each domain. 
These induced functions not only help solve the problems, but also characterize their functional distribution.

\begin{figure*}[ht]
    \centering
    \includegraphics[width=\textwidth]{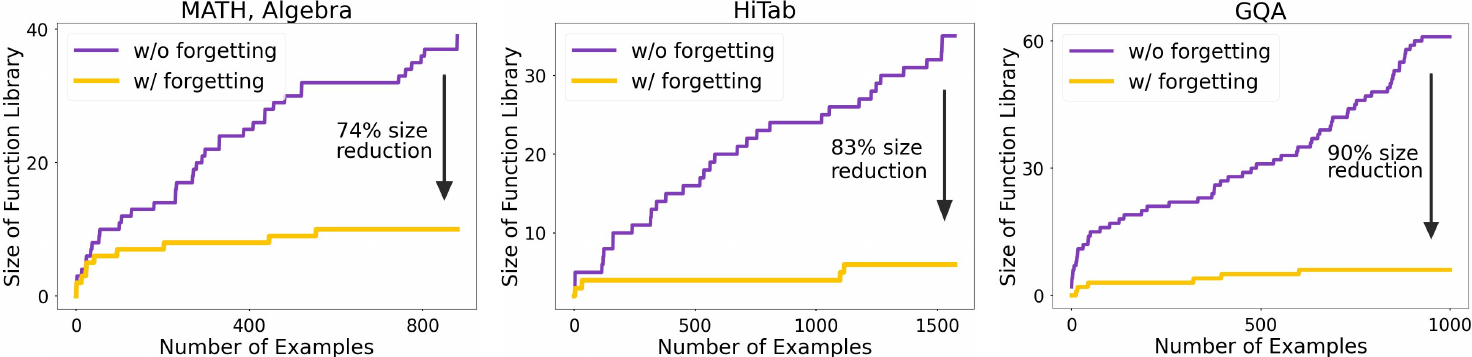}
    \vspace{-4mm}
    \caption{Library size without toolbox trimming.}
\label{fig:wo-forget}
\end{figure*}

% ##################### %
\section{Ablation Studies}
\label{sec:9:ablation-study}
% To validate our streaming setup and toolbox trimming, w
We conduct ablation studies to test the robustness to ordering (\S\ref{sub:9.1:robust-to-order}) and the importance of toolbox trimming (\S\ref{sub:9.2:wo-forget}).

\subsection{Robustness to Ordering}
\label{sub:9.1:robust-to-order}
Our method inputs examples in a streaming fashion and orders examples as they appear in the original dataset. However, it is important to study if variations in example ordering would affect final results. 
We select one dataset from each task (MATH$_{algebra}$, HiTab, GQA) as representatives for this examination.

% \paragraph{Randomized Ordering}
We shuffle the examples five times with different random seeds and run \textsc{TroVE} on each of the five orderings. We report the range of metric values and their standard deviation in \autoref{tab:example-ordering}. As a reference, we denote results (from \S\ref{sec:4:method}) using the original dataset order as \textit{original}.

\begin{table}[ht]
\small 
\centering 
\resizebox{0.42\textwidth}{!}{
    \begin{tabular}{l|r|r|r}
    \toprule
    \multirow{2}{*}{\textbf{Method / Value}} & \multicolumn{3}{c}{\textbf{Evaluation Metrics}} \\
    {} & \multicolumn{1}{c|}{acc $\uparrow$} & \multicolumn{1}{c|}{\# ops $\downarrow$} & \multicolumn{1}{c}{\# lib $\downarrow$} \\
    \midrule
    \multicolumn{4}{c}{\hlcell MATH$_{algebra}$} \\
    \midrule
    {original} & {0.25} & {18.8} & {10} \\
    \midrule
    {value range} & {0.23--0.24} & {17.3--19.0} & {5--9} \\
    {std.dev.} & {0.000} & {0.879} & {1.924} \\
    \midrule
    \midrule
    \multicolumn{4}{c}{\hlcell HiTab} \\
    \midrule
    {original} & {0.18} & {9.3} & {5} \\
    \midrule
    {value range} & {0.17--0.18} & {9.0--9.9} & {8--10} \\
    {std.dev.} & {0.003} & {0.358} & {0.837} \\
    \midrule
    \midrule
    \multicolumn{4}{c}{\hlcell \textsc{GQA}} \\
    \midrule
    {original} & {0.43} & {20.6} & {6} \\
    \midrule
    {value range} & {0.43--0.44} & {20.4--20.6} & {6--8} \\
    {std.dev.} & {0.005} & {0.150} & {0.957} \\
    \midrule
    \bottomrule
    \end{tabular}
}
\vspace{-1mm}
\caption{\textsc{CodeLLaMa} results with alternative orders.}
\vspace{-1mm}
\label{tab:example-ordering}
\end{table}

For all three datasets, no significant variance exists between the \textit{original} and \textit{randomized} ordering --- the \textit{original} results well within the \textit{randomized} value range, and standard deviations are small --- showing that \textsc{TroVE} is robust to example ordering. 
While the datasets may not be ordered in a way that optimizes function induction, the \textit{original} ordering may just be another instance of somewhat \textit{randomized} ordering.

\subsection{Without Toolbox Trimming}
\label{sub:9.2:wo-forget}

Periodic function trimming is crucial to ensure the efficiency of \textsc{TroVE}. To demonstrate this point, we compare to a \textsc{TroVE} version without toolbox trimming.
In \autoref{fig:wo-forget}, when including the trimming mechanism, the size of function libraries significantly decreases by 74\% - 90\%. The accuracy and complexity in \autoref{tab:wo-forget} also slightly degraded.

\begin{table}[ht]
\small 
\centering 
\resizebox{0.49\textwidth}{!}{
    \begin{tabular}{l|rr|rr|rr}
    \toprule
    \multirow{2}{*}{\textbf{Method}} & \multicolumn{2}{c|}{\textbf{MATH$_{algebra}$}} & \multicolumn{2}{c|}{\textbf{HiTab}} & \multicolumn{2}{c}{\textbf{GQA}} \\
    {} & \multicolumn{1}{c}{acc$\uparrow$} & \multicolumn{1}{c|}{\# ops$\downarrow$} & \multicolumn{1}{c}{acc$\uparrow$} & \multicolumn{1}{c|}{\# ops$\downarrow$} & \multicolumn{1}{c}{acc$\uparrow$} & \multicolumn{1}{c}{\# ops$\downarrow$} \\
    \midrule
    {without trim} & {0.25} & {19.6} & {0.15} & {10.5} & {0.39} & {21.1} \\
    {with trim} & {0.25} & {18.8} & {0.18} & {9.3} & {0.44} & {20.3} \\
    \bottomrule
    \end{tabular}
}
\vspace{-2mm}
\caption{\textsc{TroVE} results with and without toolbox trimming.}
\label{tab:wo-forget}
\end{table}

While the trimming threshold and time interval are easily adjustable, one can flexibly keep more functions to explore more diverse functions, or fewer due to certain constraints.
\section{Related Work}

\paragraph{Generating Program Solutions}
Many works focus on generating Python programs to solve problems, such as math \citep{ni2023lever,li2022competition,gao2023pal,chen2022program} and table QA \citep{cao2023apiassisted,cheng2023binding}. Yet most programs are built with basic operations or libraries (e.g., \texttt{sum}, \texttt{pandas}), and may be tedious and erroneous.
\citet{gupta2022visual,subramanian-etal-2023-modular,suris2023vipergpt} generate image-executable programs by hand-crafting task-specific functions, \citet{gao2023assistgpt,yang2023mm} extend this to audio and video modalities, but still require expert designs from humans.
In contrast, our work enjoys the benefit of advanced functions with reduced human labor by inducing functions using LMs.

\paragraph{Domain-Specific Library Abstraction}
\citet{shin2019program} mine common code idioms and utilize them for program synthesis. 
\citet{ellis2023dreamcoder} propose to induce functions bottom-up from a large corpus via a wake-sleep Bayesian process. \citet{pmlr-v139-wong21a} improve the search efficiency, and \citet{bowers2023top} proposed a top-down method STITCH to save memory. Most recently, LILO \citep{grand2023lilo} integrates LLMs into STITCH and abstract libraries with auto-documentation. 
While these methods all work on domain-specific logical forms, running them with general-purpose languages may vastly enlarge the search space, thus have limited applicability on many real-world tasks. 
Instead, our method generates general-purpose Python programs and can readily extend to new tasks.

\paragraph{Making Program Tools Using LLMs}
With the advances of LLMs, many works explore using LLMs to build tools.
\citet{cai2023large} examine homogenous BigBench tasks, where in each task all examples use a single tool.
\citet{qian2023creator} work on math and table QA tasks but create numerous tools that are not re-used across tasks -- this serves as our \textsc{Instance} baseline.
\citet{yuan2023craft} increase tool sharing via additional training but still yield redundant tools. 
\citet{xin2023lego} enables a growing lemma library for math theorem proving, but requires external supervision from the theorem prover and expert heuristics.
\citet{wang2023voyager} can build and learn skills in the embodied Minecraft world, yet requires self-verification and iterative refinement. 
In comparison, our method leverages execution agreement without any training or supervision.

\begin{table}[ht]
\small 
\centering 
\resizebox{0.49\textwidth}{!}{
    \begin{tabular}{l|ccccc}
    \toprule
    \textbf{\textit{programming language}} & {DreamCoder} & {PATOIS} & {STITCH} & {LILO} & {\textsc{TroVE}} \\
    \midrule
    {domain-specific} & {\cmark} & {\cmark} & {\cmark} & {\cmark} & {} \\
    {general purpose} & {} & {} & {} & {} & {\cmark} \\
    \midrule
    \textbf{\textit{tool making modules}} & {LATM} & {Creator} & {CRAFT} & {Voyager} & {\textsc{TroVE}} \\
    \midrule
    {training / curriculum} & {\cmark} & {} & {\cmark} & {\cmark} & {} \\
    {self-verification} & {\cmark} & {\cmark} & {\cmark} & {\cmark} & {} \\
    {iterative refine} & {} & {\cmark} & {\cmark} & {\cmark} & {} \\
    {self-consistency} & {} & {} & {} & {} & {\cmark} \\
    \bottomrule
    \end{tabular}
}
\vspace{-2mm}
\caption{Modules required by existing methods and our \textsc{TroVE}.}
\label{tab:compare-methods}
\end{table}

\section{Conclusion}

We proposed \textsc{TroVE}, a method for inducing a toolbox of reusable functions to use in solving programmatic tasks. 
% \textsc{TroVE} operates on the fly during test inference, and features generation with multi-mode toolbox interaction and periodic toolbox trimming.
\textsc{TroVE} produces simpler and more accurate solutions than existing methods, using sufficiently smaller function libraries. Moreover, it facilitates human program verification to be $31\%$ faster and $13\%$ more accurate. Finally, \textsc{TroVE} can induce diverse functions across tasks and datasets, shedding insights on data-specific characteristics.

\section*{Acknowledgements}

We thank Shuyan Zhou and Zhoujun Cheng for the insightful discussions about this work, Saujas Vaduguru for providing feedback about the draft, and all human participants of the verification study.


\bibliography{custom}
\bibliographystyle{icml2024}

%%%%%%%%%%%%%%%%%%%%%%%%%%%%%%%%%%%%%%%%%%%%%%%%%%%%%%%%%%%%%%%%%%%%%%%%%%%%%%%
%%%%%%%%%%%%%%%%%%%%%%%%%%%%%%%%%%%%%%%%%%%%%%%%%%%%%%%%%%%%%%%%%%%%%%%%%%%%%%%
% APPENDIX
%%%%%%%%%%%%%%%%%%%%%%%%%%%%%%%%%%%%%%%%%%%%%%%%%%%%%%%%%%%%%%%%%%%%%%%%%%%%%%%
%%%%%%%%%%%%%%%%%%%%%%%%%%%%%%%%%%%%%%%%%%%%%%%%%%%%%%%%%%%%%%%%%%%%%%%%%%%%%%%
\newpage
\appendix
\onecolumn

\section{Prompt Example}
\label{app:a:prompt-example}

We introduced baseline methods (\textsc{Primitive} and \textsc{Instance}) in \S\ref{sub:3.2:baseline-methods} and our main method in \S\ref{sec:4:method}. To more concretely illustrate the prompts beyond textual description, we provide figure examples.

\begin{figure*}[ht]
    \centering
    \includegraphics[width=0.6\textwidth]{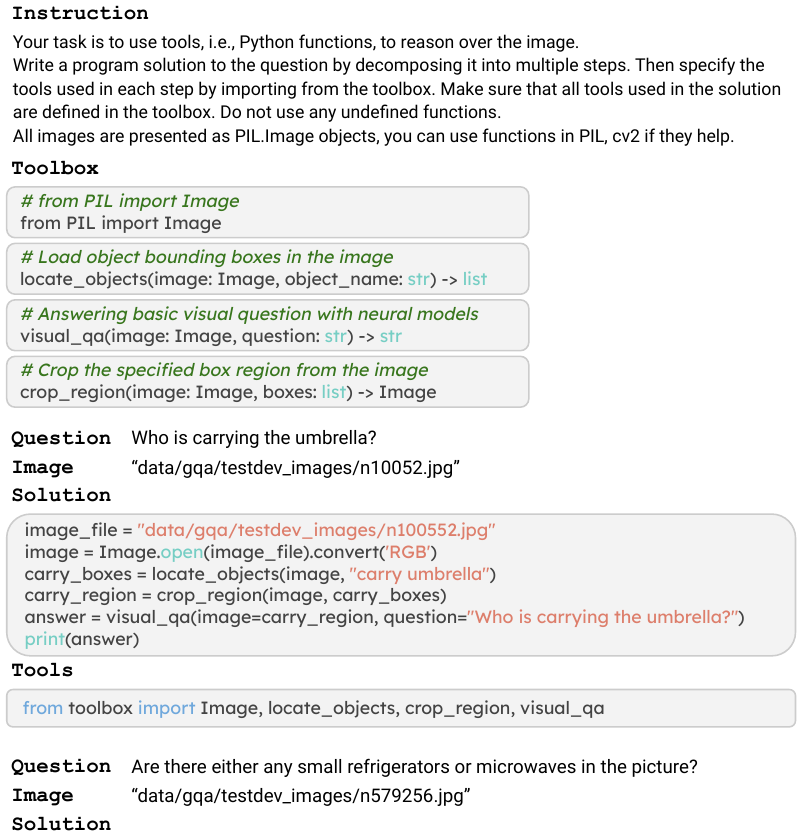}
    \caption{Example prompt in the primitive setting.}
\label{fig:prompt-primitive}
\end{figure*}

\autoref{fig:prompt-primitive} is an example prompt used in the \textsc{Primitive} setting, where on the bottom is the current test example and \textit{Solution} is expected to be filled by the model.

We use the same format in the \textsc{Instance} setting, but changed the \textit{Instruction} to be: 
\begin{mdframed}
{\small Your task is to write the solution with high-level tools, i.e., Python functions, to reason over the image.

Think about the potential program solution for this example, you can create high-level functions, that could be used to solve this example. For example, if the solution involves multiple actions that are always used together, it is more efficient to create and use the tool.}
\end{mdframed}

Similarly, in the \textsc{Online} setting, respectively for the \textsc{create}, \textsc{import}, and \textsc{skip} modes, the \textit{Instruction}s reads:
\begin{mdframed}
{\small Your task is to write Python program solutions to reason over images.
You should also create Python functions that can be used by your solution, if you believe the function can be reused to solve other questions.
}
\end{mdframed}

\begin{mdframed}
{\small Your task is to write Python program solutions to reason over images.
The toolbox section lists all the available functions that can be used in your solution.
}
\end{mdframed}

\begin{mdframed}
{\small Your task is to write Python program solutions to reason over images.}
\end{mdframed}
\section{Evaluation Metric Details}
\label{app:b:metric-detail}

\paragraph{Answer Correctness}
We measure if the solution execution result matches the annotated answers. For answers that are expected in a textual format (e.g., ``brown''), we use the Exact Match (EM) metric; for numerical answers, we convert it to float type and round to two decimals, then measure if the values match (with a difference less than $1\mathrm{e}{-6}$).

\paragraph{Program Complexity}
We quantify the complexity of programs in their number of operations.
Concretely, we separate solutions into multiple expressions, and parse each expression into abstract syntax trees (AST). We take the depth of each AST as the number of operations conducted in the corresponding expression. We sum this value of all expressions and denote it as the complexity (i.e., number of operations) of the entire program.

\section{Human Study: Individual Results}
\label{app:c:human-study}

In the verification human study, performance between people may vary due to their programming expertise. Therefore, in this section, we present more detailed results of individual participants, and justify the significance of our findings.

\begin{figure}[ht]
\centering
    \includegraphics[width=0.80\textwidth]{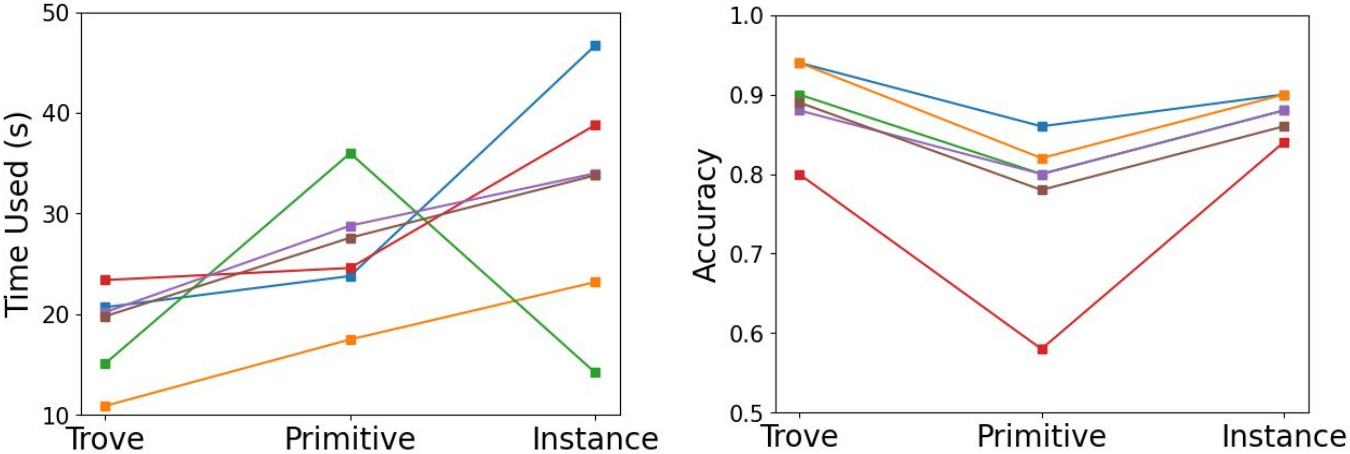}
    \caption{Individual verification accuracy and time used.}
\label{fig:individual-verify}
\end{figure}

As shown in \autoref{fig:individual-verify} (left), verification accuracy is higher for tool-involved methods (\textsc{TroVE} and \textsc{Instance}) compared to using primitive functions only (\textsc{Primitive}). Most people perform verification more accurately on programs produced by \textsc{TroVE} except one person (the red line), which also has lower detection accuracy in general.

Their times used for verification are shown on the right, with the same line color identifying each human. Most people find \textsc{TroVE} the fastest, \textsc{Primitive} taking $8.2$ more seconds on average, and \textsc{Instance} requires another $10.8$ seconds. However, one person responded differently and personally found \textsc{Instance} more efficient than \textsc{Primitive}, which is the major contribution of the relatively large variance of \textsc{Instance} reported in \S\ref{sec:7:human-verify}.

\end{document}